# Human Detection for Night Surveillance using Adaptive Background Subtracted Image

**Yash Khandhediya**
Electronics and Communication Department, L. D.
College of Engineering
*khandhediya.yash.364@ldce.ac.in*

**Karishma Sav**
Electronics and Communication Department, L. D.
College of Engineering
*karishma.naresh.51@ldce.ac.in*

**Vandit Gajjar**
Electronics and Communication Department, L. D.
College of Engineering
*gajjar.vandit.381@ldce.ac.in*

*Abstract* – Surveillance based on Computer Vision has become a major necessity in current era. Most of the surveillance systems operate on visible light imaging, but performance based on visible light imaging is limited due to some factors like variation in light intensity during the daytime. The matter of concern lies in the need for processing images in low light, such as in the need of nighttime surveillance. In this paper, we have proposed a novel approach for human detection using FLIR (Forward Looking Infrared) camera. As the principle involves sensing based on thermal radiation in the Near IR Region, it is possible to detect Humans from an image captured using a FLIR camera even in low light. The proposed method for human detection involves processing of Thermal images by using HOG (Histogram of Oriented Gradients) feature extraction technique along with some enhancements. The principle of the proposed technique lies in an adaptive background subtraction algorithm, which works in association with the HOG technique. By means of this method, we are able to reduce execution time, precision and some other parameters, which result in improvement of overall accuracy of the human detection system.

*Keywords*—Adaptive Background Subtraction, Human Detection, Thermal, FLIR, Nighttime Surveillance, Dynamic Image Processing

## I. INTRODUCTION

The contemporary human detection systems are used in applications like Autonomous Vehicles, Headcounters, Search and Rescue Operations, etc. but these surveillance systems limit itself in night surveillance due to the use of RGB cameras. As we know that, there are myriads of applications like border surveillance, security purposes, monitoring systems, anomaly or intruder detection, which most seek a system capable of night surveillance.

Many researchers today have proposed efficient methods of detecting humans from images, the most common area of research being pedestrian detection. According to a research done in 2015, researchers proposed many techniques; at times a fusion of two or more; for the purpose of feature extraction from an image. This research also showed that the lowest miss rate achieved at that time was of 22.49 % [1]. However, there are a very few techniques proposed for nighttime surveillance. In most of the systems proposed for Nighttime surveillance for human detection, Thermal Imaging is the most frequently used method, where the features required for Human detection are extracted from thermal images. Thermal imaging uses NIR (Near Infrared) band of Infrared light, which ranges from (0.75 to 1.4 μm). Some of the existing methodologies make use of thermal cameras along with visible light cameras [2], by the means of image fusion but that requires twice the amount of hardware. On the other hand, some directly perform the feature extraction on the thermal images reducing the hardware but at the same time compromising with accuracy.

The aim of this paper is to propose a technique, which provides better accuracy and precision, while minimizing the hardware and execution time requirements.

## II. NEED OF PROPOSED METHOD

The conventional methods of Human Detection use RGB imaging as input. However, for the application of Human Detection in Nighttime Surveillance, we cannot use RGB imaging as the input to our system because at nighttime the amount of light intensity available is inadequate to produce clear images. As a solution to this problem, the proposed method makes use of an FLIR camera and processes on thermal images; which produces clear images irrespective of the amount of light intensity.

The most widely used feature extraction technique is HOG, Histogram of Oriented Gradients[3]. In spite of good performance of HOG, there also exist some limitations. Firstly, using HOG in real time application is not feasible due to large execution time. Secondly, the output of HOG has high recall but low precision of this output is a matter of concern. So we always have to use some extra filtering or feature extracting technique to enhance the performance of the HOG.

Another technique, which is used in for Human Detection, is background subtraction from current image. This technique can be used to enhance the performance of HOG, as only background subtraction has the capability to detect anomalies in a scene. But this technique is currently only applicable to images from a static camera. On the other hand, the proposed technique will be applicable to images from dynamic cameras. As proposed in our technique, a fused image is created using the original image and the background subtracted image and the performance of HOG on fused image improves significantly as compared to when applied on the original image.

Another improvement proposed by our technique is adaptive background subtraction for dynamic camera. We have used FLIR camera in capturing images. FLIR cameras do not provide a wide-angle view. We can use a number of cameras to solve this problem but it will not be cost effective and the system would become more complex as it would also need synchronization between all the cameras. Therefore, to increase the span, the camera is made to rotate to cover a broader area.

### III. PROPOSED METHOD

The proposed method reduces the execution time taken by HOG technique by means of Background Subtraction. In the block diagram shown in Fig. 1, round blocks represent our signals and rectangular blocks represent functional blocks. Background Image, Input Image, Background subtracted image are signals and Dynamic background adaption block, Image Fusion Block, HOG feature extraction block and linearly trained SVM block are functional bocks. We will describe the algorithm proposed by us for two different cases.

#### A. Static Camera: Constant Background Subtraction

For simplicity, first we have applied our algorithm on thermal images taken from a static camera. The source for these images is OTCBVS Benchmark Dataset Collection [4]. In this case we have a single background image which is subtracted from every image. After background subtraction, the original image and the background subtracted image are fused together; and then HOG features are extracted from this fused image.

The output of the HOG Feature Extraction block is then fed to Linearly Trained SVM, which generates the prediction Matrix. The prediction matrix when multiplied with the input image provides us with the desired output. The results for this algorithm are summarised in Fig. 2.

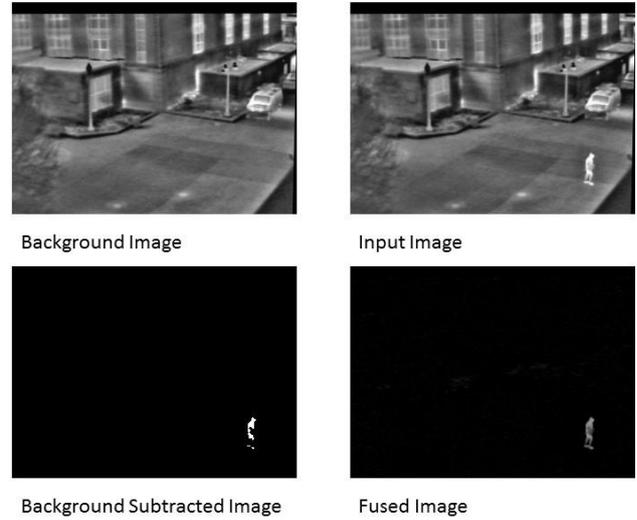

Figure 2 Results for Static Camera

#### B. Dynamic Camera: Adaptive Background Subtraction

As we are using FLIR Cameras, we need the camera to be moving to cover a wider angle. This requires an altogether modified technique. Therefore, we need to modify the algorithm, to work with dynamic cameras with. The part of the block diagram in Fig. 1 shown in Dotted lines is included to accommodate the input from a dynamic camera. The purpose of this paper lies in the design of the algorithm which accommodates a dynamically changing background. This algorithm corresponds to the proposed feature extraction technique which will work effectively with dynamic camera images. The working of this algorithm is summarised in the Block Diagram in Fig. 3. However, the detailed explanation and working of the algorithm is presented in the Next Section. The images for experimentation are taken from ETHZ Thermal Infrared Dataset [5].

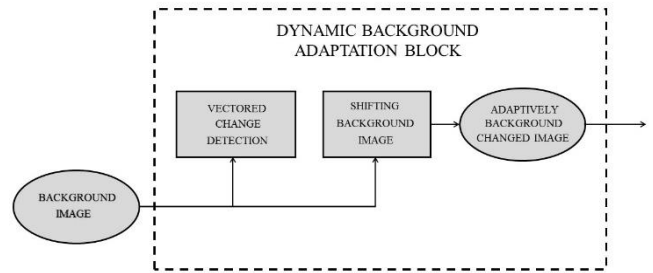

Figure 3 Dynamic Background Adaptation Block

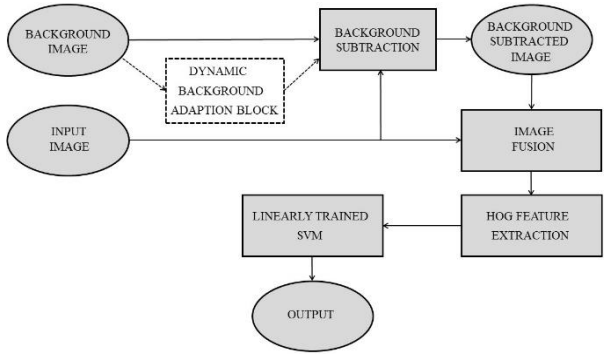

Figure 1 : Block Diagram for Proposed Algorithm

## IV. THE PROPOSED FEATURE : ADAPTIVE BACKGROUND SUBTRACTION

The working principle of this feature is very simple yet effective. As the problem arises due to a dynamic camera whose background changes, the algorithm is designed to shift a constant background by an amount, which is in proportion with the rotation of the camera. The precise working of this algorithm can be subdivided into two parts for detailed analysis.

### A. Vectored Change Detection

In this part of the algorithm, we are firstly selecting a particular region of the image, which has a specific object very distinct in properties from its neighboring pixels; i.e.; an object which does not blend into the background. Also the object under consideration should be static with respect to the background. To demonstrate this logic, let us consider the image shown in Fig. 4.

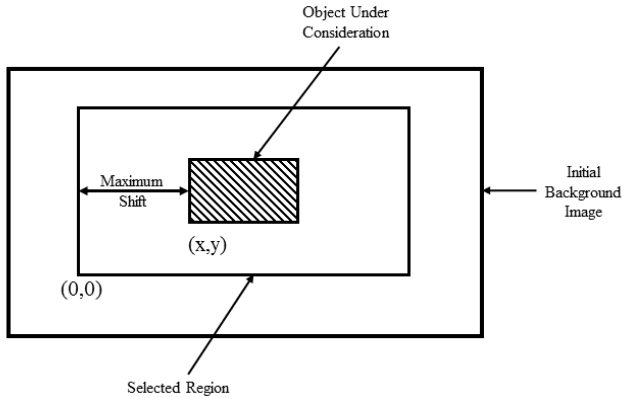

Figure 4 Region and Object under Consideration for Initial Background

The outermost rectangle represents our image. The innermost rectangle shaded with lines represents the object under consideration, while the rectangle exterior to the object under consideration represents our selected region. The region is so selected that initially the object under consideration is exactly at its center. Let the coordinates of the object under consideration with respect to the selected region be (x, y).

Let us now consider a second image, which can be an input image for any arbitrary angle of rotation. The Selected Region will remain same. Now the selected region in this second input image will be scanned for the Object under consideration. This will be done by taking a window frame of the size of the object and moving this frame in the entire Selected region. The error is calculated for each frame and the window frame which has the minimum error now becomes the new location of our Object. To visualize the shift in the new image, consider Fig. 5. Let the location of the object under consideration in the input image be (x', y'). If we compare the two location coordinates of both the images, we can get a difference vector which can be utilized to determine the shift required in the background.

$\Delta x = x' - x$     is the change in x coordinate;

$\Delta y = y' - y$     is the change in y coordinate;

($\Delta x$, $\Delta y$) represent the difference vector coordinates, where the magnitude and angle of the vector are as described in the equations as follows:

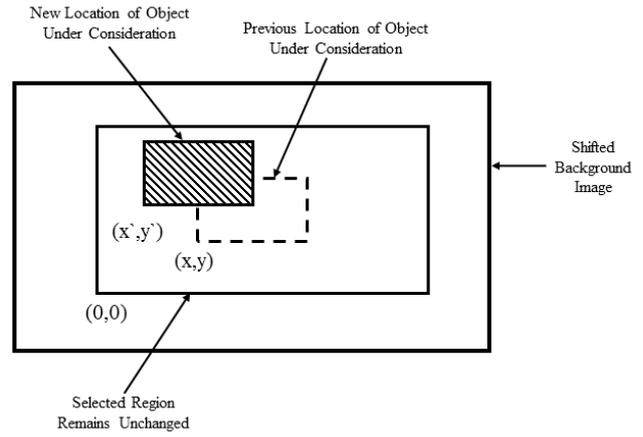

Figure 5 Object under Consideration for Shifted Camera Image

$|\Delta| = \sqrt{\Delta x^2 + \Delta y^2}$     is the magnitude of the difference vector;

$Arg(\Delta) = \tan^{-1}(\Delta y / \Delta x)$     is the angle of the difference vector.

Therefore, we now have a difference vector in terms of magnitude and angle. The magnitude of the vector represents the amount of shift while the angle represents the direction of shift.

### B. Shifting Background Image

Now we know the magnitude and direction of shift of a static object under consideration. Because it is a static object, i.e. it does not move with respect to the camera, the difference vector actually indicates a shift or movement of the camera. As the object is static with respect to the background, the shift of the background has to be same as the shift in the object. Therefore, if we shift the original background by this difference vector we will have a shifted background. This phenomenon is clearly depicted in Fig. 6.

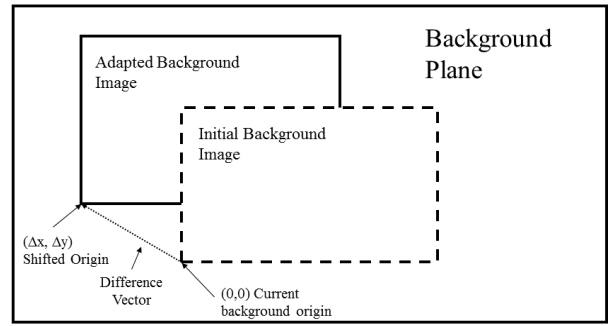

Figure 6 Adapting Background to Dynamic Camera

Using the procedure explained above an algorithm is created which can create Adaptive Backgrounds for Subtraction from images taken from a Dynamic Camera. An image taken from a dynamic camera is processed using both constant backround subtraction and Adaptive background subtraction and the output and original image are as shown in Fig. 7.

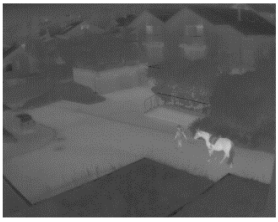
Original Image

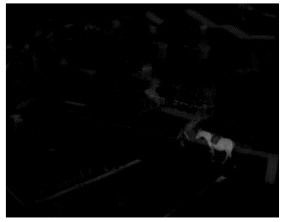
Background Subtracted Image

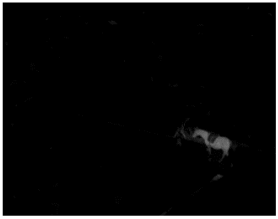
Adaptive Background Subtracted Image

Figure 7 Output Comparison between images processed using constant background subtraction and Adaptive Background Subtraction

It is clearly visible that the proposed technique is more efficient in extracting features as compared to the available technique.

Table 1 Comparison of HOG, HOG with Background Subtraction and HOG with Adaptive Background Subtraction for Static and Dyamic Cameras

| Parameter | HOG | HOG+BS for Static Camera | HOG + BS for Dynamic Camera | HOG+ABS for Dynamic Camera |
|---|---|---|---|---|
| Execution Time ( in Seconds) | 1768.29 | 23.0047 | 631.0068 | 478.0096 |
| Precision | 12.36% | 83.11% | 58.36% | 76.09% |
| Recall | 100% | 100% | 100% | 100% |

## V. RESULTS & COMPARISON

For Comparison purposes, we have first detected humans using only HOG feature Extraction and the results are compared when HOG is used in association with Background Subtraction and also with the results obtained when HOG is used along with our proposed technique.

Various significant parameters are compared in Table 1. The Recall is 100% in all the techniques, but we can see that there is a drastic improvement in Precision. Also the execution time is significantly reduced. The execution time increases if we use a dynamic camera and the precision also reduces but given the flexibility for movement and coverage of greater angle of the scene is a feasible trade-off.

## VI. CONCLUSION

In this paper, we have proposed a nighttime surveillance system based on processing of thermal images taken from a FLIR camera. In addition to this, we have proposed a novel technique, named Adaptive Background Subtraction, for feature extraction from images taken from a dynamic camera. The proposed feature has been true to its necessity as it catered to the system by increasing precision, allowing the camera some freedom of movement and reducing the execution time of the system.

## VII. ACKNOWLEDGEMENT

We would like to express our gratitude to Prof. Usha Neelkantan, HOD of Electronics and Communication Department. L. D. College of Engineering for her constant support and motivation. We would also like to thank Prof. Kinnar Vaghela for without his guidance we might not have been able to complete our research successfully.

## VIII. REFERENCES

[1] Benenson R., Omran M., Hosang J., Schiele B. (2015) Ten Years of Pedestrian Detection, What Have We Learned?. In: Agapito L., Bronstein M., Rother C. (eds) Computer Vision - ECCV 2014 Workshops. ECCV 2014.

[2] Alex Leykin, Yang ran and Riad Hammoud, "Thermal-Visible Video Fusion for Moving Target Tracking and Pedestrian Classification" In: Computer Vision and Pattern Recognition, 2007. CVPR '07.

[3] Shyang-Lih Chang, "Nighttime pedestrian detection using thermal imaging based on HOG feature" In: 2011 International Conference on System Science and Engineering.

[4] IEEE OTCBVS WS Series Bench Zhang, M.M, Choi, J., Daniilidis, K., Wolf, M.T. & Kanan, C. (2015) VAIS: A Dataset for Recognizing Maritime Imagery in the Visible and Infrared Spectrums. In: Proc of the 11th IEEE Workshop on Perception Beyond the Visible Spectrum

[4] (PBVS-2015).

[5] ETHZ Thermal Infrared Dataset, ASL Datasets